\documentclass{bmvc2k}

\usepackage{booktabs}
\usepackage{times}
\usepackage{soul}
\usepackage{geometry}
\usepackage{url}
\usepackage[utf8]{inputenc}
\usepackage[small]{caption}
\usepackage{graphicx}
\usepackage{amsmath}
\usepackage{amsthm}
\usepackage{amssymb}
\usepackage{amsfonts}
\usepackage{booktabs}
\usepackage{algorithm}
\usepackage{algorithmic}
\usepackage[switch]{lineno}
\usepackage{xcolor}
\usepackage{pifont}
\usepackage{hyperref}

\definecolor{darkgreen}{rgb}{0.0, 0.6, 0.0} 

\pagecolor{white} 

\newcommand{\xmark}{\ding{55}} 

\newcommand{\tablesmall}{\fontsize{9}{8.4}\selectfont}

\title{Few-Shot Classification of Interactive Activities of Daily Living (InteractADL)}

\addauthor{Zane Durante}{}{1}
\addauthor{Robathan Harries}{}{1}
\addauthor{Edward Vendrow}{}{1}
\addauthor{Zelun Luo}{}{1}
\addauthor{Yuta Kyuragi}{}{2}
\addauthor{Kazuki Kozuka}{}{2}
\addauthor{Li Fei-Fei}{}{1}
\addauthor{Ehsan Adeli}{eadeli@stanford.edu}{1}

\addinstitution{Stanford University}
\addinstitution{Panasonic Corporation}

\runninghead{Durante et al.}{Interactive Activities of Daily Living (InteractADL)}

\def\etal{\emph{et al}\bmvaOneDot}

\begin{document}

\newgeometry{left=.75in, right=.25in, top=.5in, bottom=.25in} 

\maketitle

\begin{abstract}
Understanding Activities of Daily Living (ADLs) is a crucial step for different applications including assistive robots, smart homes, and healthcare. However, to date, few benchmarks and methods have focused on complex ADLs, especially those involving multi-person interactions in home environments. In this paper, we propose a new dataset and benchmark, InteractADL, for understanding complex ADLs that involve interaction between humans (and objects). Furthermore, complex ADLs occurring in home environments comprise a challenging \textit{long-tailed distribution} due to the rarity of multi-person interactions, and pose \textit{fine-grained} visual recognition tasks due to the presence of semantically and visually similar classes.  To address these issues, we propose a novel method for fine-grained few-shot video classification called \textit{Name Tuning} that enables greater semantic separability by learning optimal class name vectors. We show that Name Tuning can be combined with existing prompt tuning strategies to learn the entire input text (rather than only learning the prompt or class names) and demonstrate improved performance for few-shot classification on InteractADL and 4 other fine-grained visual classification benchmarks.  For transparency and reproducibility, we release our code \href{https://github.com/zanedurante/vlm_benchmark}{here}.
\end{abstract}

\restoregeometry

\section{Introduction}

\begin{figure}[t]
    \centering
    \includegraphics[width=0.88\linewidth]
    {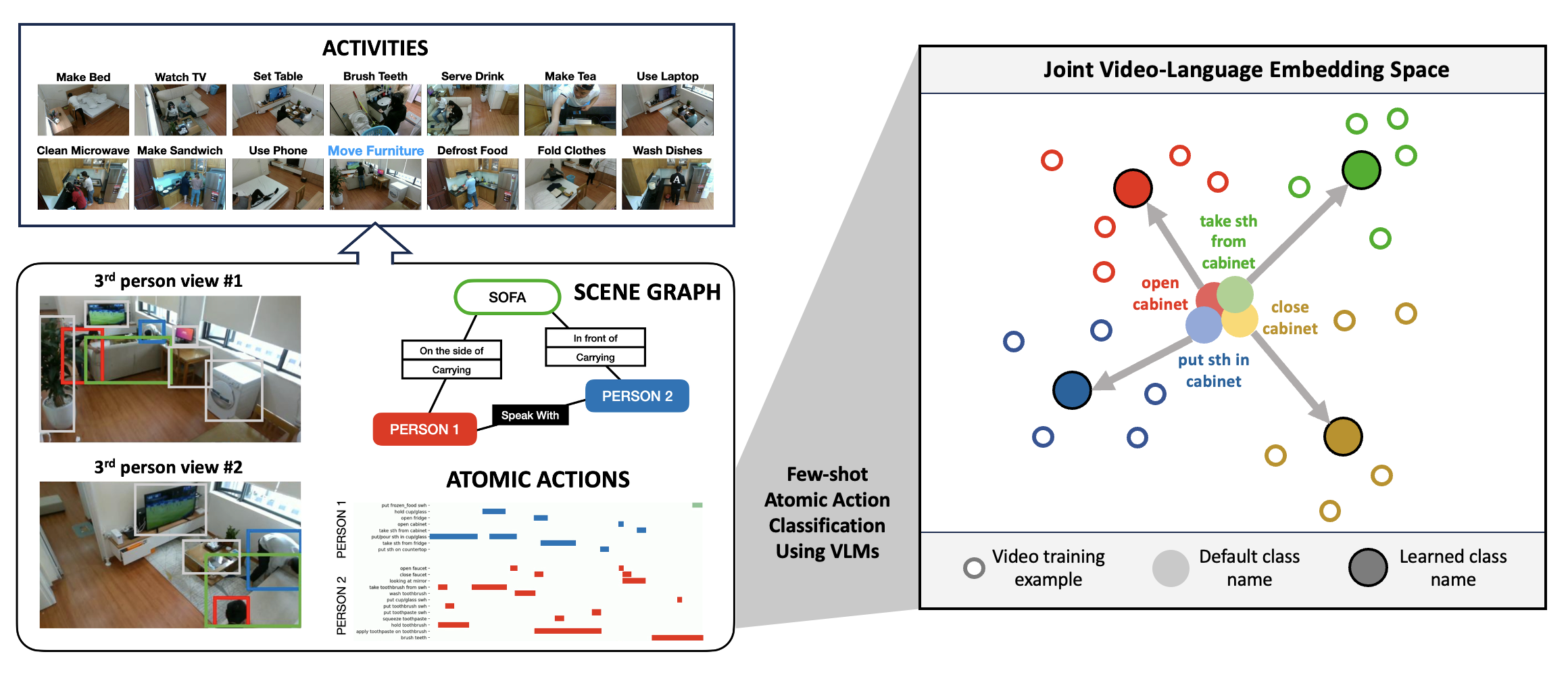}
    \vspace{1mm}
    \caption{Our fine-grained visual recognition dataset InteractADL poses a challenging classification task for Visual Language Models (VLMs). Left: a sample timeline of multiple ADLs in our InteractADL dataset. We show two $3^{\text{rd}}$ person views along with annotated higher-level activity labels, temporal atomic actions, and dense spatiotemporal scene graphs. Right: dual encoder VLMs have a joint video-language embedding space, in which similarity scores between an input video and category names are computed. For these VLMs, we seek to use a small number of training examples to learn better class names that provide greater semantic separation and improved classification performance.}
    \label{fig:combined_fig1}
\vspace{-3mm}
\end{figure}

Activities of Daily Living (ADLs) are defined as a set of basic, essential, and routine tasks that most healthy people can perform without any assistance \cite{katz1983assessing}. 
Recognizing ADLs is an essential component in several applications, from smart homes \cite{debes2016monitoring} and assistive robots \cite{lauretti2017learning} to automated healthcare assessment \cite{desai2004activities} applications. 
However, there are few benchmarks and methods for understanding human ADLs at home, especially where there is interaction with other humans. Prior work has focused on ADLs at home with a single person present, such as Toyota Smart Home \cite{dai2022toyota} and Home Action Genome (HOMAGE) \cite{rai2021home}, or benchmark ego-view scenes in specific rooms in the house, e.g., in the kitchen as in EPIC-Kitchens \cite{damen2018scaling}. 

There are several challenges associated with understanding ADLs at home. Specifically, (1) such activities are often long and sometimes take hours (e.g., house cleaning). They are often composed of several atomic actions (indivisible sequences of primitive actions that are completed without interruption and often take 2-3 seconds); (2) In addition to the relationships between humans and objects, it is crucial to model interactions between multiple humans when building models for recognizing ADLs. Specific interactions between humans are often rare, forming a long-tailed distribution that poses challenges for large-scale training; (3) ADLs consist of many classes that are visually and semantically similar, posing a fine-grained action recognition task that is challenging for many current methods. 

In this paper, we introduce a new dataset and benchmark for complex video understanding of human interactions and ADLs in the home environment called \textit{InteractADL}. The dataset includes RGB videos from multiple views, including each of two people's ego-view and between 2-19 $3^{\text{rd}}$ person views per video. In addition to the object and human bounding boxes annotated in the dataset, it includes dense annotation of human-object (H-O) relationships and interactions between humans (H-H) in the form of spatiotemporal scene graphs. The sequences are further annotated with activity labels at two levels of abstraction, i.e., overall activity class and atomic action segments (see Fig.~\ref{fig:combined_fig1}). The tasks for this benchmark comprise detecting dynamic H-H interactions as well as few-shot detection of higher-level activities and atomic actions in the videos.


Inspired by our new dataset and benchmark, we also propose a novel method for fine-tuning visual-language models (VLMs).  Our method, called \textit{Name Tuning}, enables greater semantic flexibility and separability between category names by allowing them to be fine-tuned for fine-grained recognition tasks. We show that Name Tuning achieves superior performance on InteractADL compared to existing dual encoder fine-tuning methods and show that Name Tuning achieves state-of-the-art performance across 3 separate few-shot video classification benchmarks and can perform on par with existing methods without using any video pre-training data.

\section{Related Work}

\begin{table*}[t]
    \label{tab:adl_comparison}    
    \centering
    {\tablesmall
    \setlength\tabcolsep{2.0pt}
    \begin{tabular}{lcccccccccccccc}\toprule
                & &            & Activity & Atomic & &  & 
                         Multi-  & Multi- & View  \\
        Activity Datasets & ADL & Hours &  Classes & Classes&  HOI & HHI &
                         View  & person & Type  \\
     \midrule
   
ToyotaTrimmed \cite{das2019toyota} & \checkmark & 53.1 & 5 & 31     & \xmark & \xmark & \checkmark & \xmark & M \\
TSU \cite{dai2022toyota} & \checkmark & 188 & 5 & 51             & \xmark & \xmark & \checkmark & \xmark & M \\
CAD-120 & \checkmark & 0.5  & 10 & 20            & \xmark & \xmark & \xmark & \xmark & S \\
DAHLIA \cite{vaquette2017daily} & \checkmark & 33.4 & 7 & -                & \xmark & \xmark & \checkmark & \xmark & M \\
PKU-MMD \cite{liu2017pku} & \checkmark & 49  & - & 51         & \xmark & \xmark & \checkmark & \checkmark & S \\
Charades \cite{sigurdsson2016hollywood}  & \checkmark & 82  & - & 157         & \checkmark & \xmark & \xmark & \checkmark & M \\
Charades-ego \cite{sigurdsson2018charades} & \checkmark & 68.8  & - & 157    & \checkmark & \xmark & \xmark & - & E \\
ETRI-Activity3D \cite{jang2020etri} & \checkmark & -  & - & 55  & \xmark & \xmark & \checkmark & \xmark & S \\
MSR-DA3D \cite{wang2012mining} & \checkmark   & 16 & - & -     & \xmark & \xmark & \xmark & \xmark & S \\  \hline 
\rule{0pt}{2ex}RGBD-HuDaAct \cite{ni2011rgbd} & \xmark & 46  & - & 12  & \checkmark & \xmark & \xmark & \xmark & S \\

ActivityNet \cite{caba2015activitynet} & \xmark & 849  & - & 200  & \checkmark & \xmark & \xmark & \xmark & S \\

Kinetics-700 \cite{carreira2019short} & \xmark & 1167  & - & 700 & \checkmark & \xmark & \xmark & \xmark & S \\

AVA \cite{gu2018ava} & \xmark & 107.5  & - & 80 & \checkmark & \checkmark & \xmark & \checkmark & S \\

EPIC-Kitchens \cite{damen2018scaling} & \xmark & 55  & - & - & \checkmark & \xmark & \xmark & \xmark & E \\

MMAct \cite{kong2019mmact} & \xmark & 126.7  & - & 37 & \checkmark & \xmark & \checkmark & \xmark & S \\

ActionGenome \cite{ji2020action} & \xmark & 82  & - & 157  & \checkmark & \xmark & \xmark & \checkmark & M \\

Breakfast \cite{kuehne2014language} & \xmark & 77 & - & 10 & \checkmark & \xmark & \checkmark & \xmark & S \\

Ego4D \cite{grauman2021ego4d} & \xmark & 3670 & - & - & \checkmark & \xmark & \xmark & \xmark & E \\

HOMAGE \cite{rai2021home} & \xmark & 25.4  & - & - & \checkmark & \xmark & \checkmark & \xmark & E + S \\ \hline
\rule{0pt}{2ex}\textbf{InteractADL} & \checkmark & \textbf{242}  & \textbf{50} & \textbf{334}  & \textbf{\checkmark} & \textbf{\checkmark} & \textbf{\checkmark} & \textbf{\checkmark} &  M + T + E\\

    \bottomrule  \end{tabular}}
    \vspace{2.5mm}
    \caption{Comparison with other ADL (Activities of Daily Living) Datasets (top) and related video datasets (bottom). HHI: Human-human interaction, HOI: Human-object interaction. View types: M: Monitoring, S: Shooting, E: Ego, T: Top-Down.}
    \label{tab:dataset_table}
    \vspace{-3mm}
\end{table*}

\textbf{Data Acquisition of ADLs at Home.}  Prior work in building datasets of human actions has varied in their degree of spontaneity, with some designs requiring participants to act out specific action classes and steps, such as PKU-MMD \cite{liu2017pku}, Charades \cite{sigurdsson2018charades}, ETRI-activity3D \cite{jang2020etri}, Rgbd-hudaact \cite{ni2011rgbd}, MMAct \cite{kong2019mmact}, and ActionGenome \cite{ji2020action}. In contrast, some other datasets have given more agency to actors with unscripted actions, including Toyota \cite{das2019toyota}, DAHLIA \cite{vaquette2017daily}, and Ego4D \cite{grauman2021ego4d}. 

Capturing multiple viewpoints can allow models to become more robust to variations of input. A few prior works have, accordingly, focused on acquiring multiple viewpoints, with \cite{das2019toyota} utilizing 7 cameras, \cite{vaquette2017daily} utilizing 3 Kinect v2 sensors, \cite{kong2019mmact} utilizing 4 camera views and 1 egocentric view, and \cite{kuehne2014language} utilizing 3 to 5 views depending on the location.

Previous data acquisition methods from home environments often specialize in single-person actions, like Epic-Kitchens \cite{damen2018scaling}. 
Additionally, prior work has also focused on human-object interaction, namely JRDB-Act \cite{ehsanpour2021jrdb} and ActionGenome \cite{ji2020action}. 
However, actions in the real world often involve multiple individuals interacting with each other and objects. Few datasets involve multiple individuals for a given task, and such configurations tend to comprise only a portion of the dataset \cite{liu2017pku,ji2020action}. 
In contrast to previous datasets, InteractADL captures spontaneous ADLs between two individuals and provides high-level action, atomic action, and spatiotemporal scene graph annotations to capture human-human interactions while providing both ego and $3^\text{rd}$ person views. We highlight our explicit differences with existing datasets in Table \ref{tab:dataset_table}.


\noindent \textbf{Large Pre-trained Visual-language Models.}
An increasingly popular paradigm when training visual-language systems is to leverage vast amounts of task-agnostic general internet data and use self-supervised learning to train large visual-language models (VLMs) that perform well on a variety of downstream tasks \cite{ImageCLIP,alayrac2022flamingo}.  Using matching visual-text pairs from the internet, such models can be trained with a variety of self-supervised objectives, including contrastive learning or image captioning \cite{yu2022coca}. By modeling computer vision tasks using natural language, these VLMs tend to exhibit powerful zero-shot and few-shot capabilities over a variety of domains \cite{Coop-zhou2021learning,alayrac2022flamingo}. 

\begin{figure}[t]
    \centering
    \includegraphics[width=0.80\linewidth]{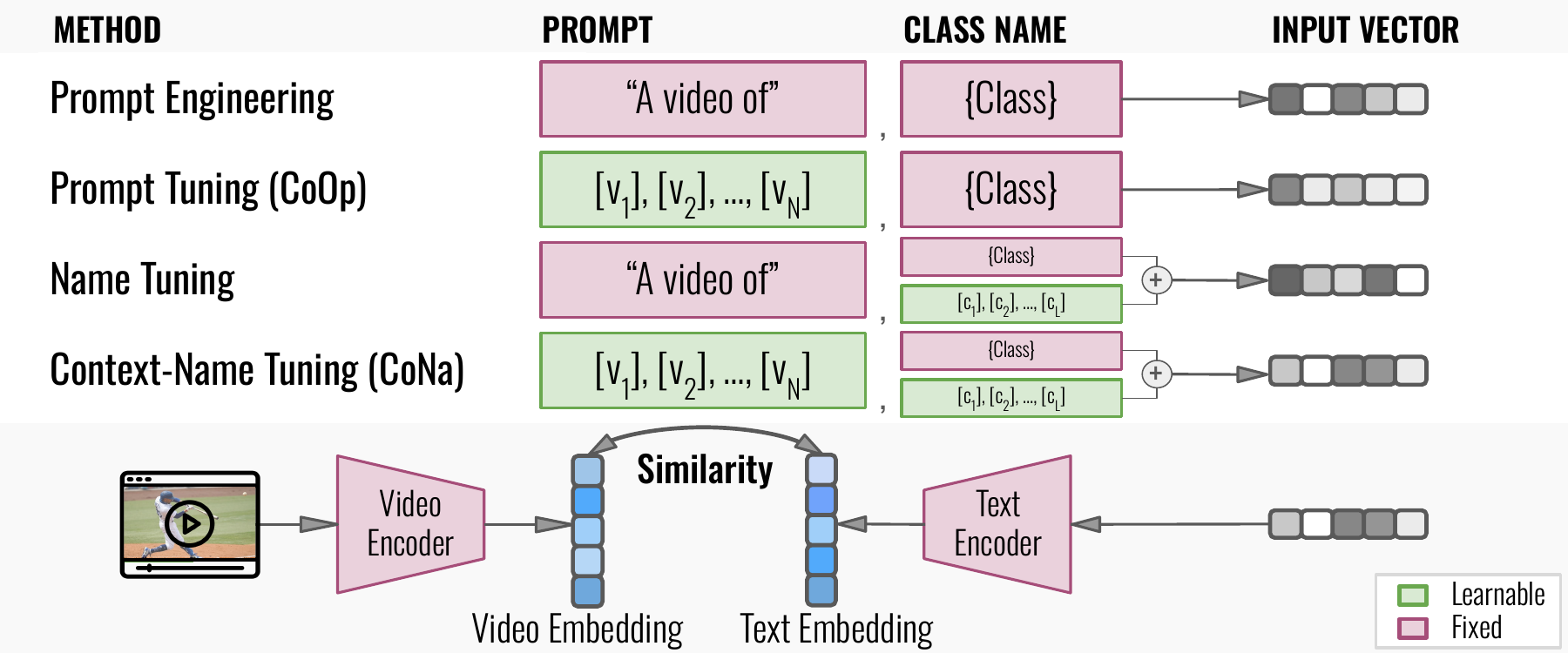}\
    \vspace{2mm}
   \caption{Comparison of input-text optimization methods for dual encoder VLMs. We contrast the methods presented in this work, Name Tuning and CoNa, with standard prompt engineering and prompt tuning (CoOp). We introduce learnable offset vectors to fine-tune class names in the input text.}
    \label{fig:model_fig}
\end{figure}

\vspace{4mm}
\noindent \textbf{Fine-tuning Large Pre-trained Models.}
Fine-tuning pre-trained large language models (LLMs) for downstream tasks has been a very popular approach in Natural Language Processing (NLP) \cite{he2021towards,liu2021promptingmethods}. One approach for tuning large language models uses prompt-tuning methods to train token embeddings that act as a prompt alongside additional text input \cite{li2021prefixtuning,liu2021promptingmethods}. Versions of prompt-tuning \cite{Coop-zhou2021learning,lu2022prompt,zhou2022conditional} have been applied to CLIP, allowing efficient fine-tuning for image classification and text-image retrieval. We note that these existing methods use class names as fixed, unlearnable tokens within class-specific prompts, whereas our work demonstrates the potential advantages of fine-tuning these class name tokens.  We highlight the difference between our method and existing prompt tuning methods in Fig. \ref{fig:model_fig}.

\vspace{-2mm}
\section{InteractADL}

InteractADL is a multi-view RGB video dataset that provides detailed annotations for understanding daily living  activities and interactions in a home environment. The dataset is recorded on sensors capturing video recordings synchronized across multiple viewpoints (ego-view and $3^{\text{rd}}$ person). Given the difficulty of obtaining videos of home environments, this dataset provides an ego-view and $3^{\text{rd}}$ person views for all videos. InteractADL provides hierarchical action labels with high-level activity and low-level action information, human and object bounding boxes, and dense human-human and human-object interaction labels.

\vspace{-1mm}
\subsection{Annotation Protocol}

For each task performed in a video, pairs of participants were selected such that the same pairs were not used for different tasks. Vidat \cite{zhang2020vidat} was used as a tool to annotate activities and atomic actions per frame using object bounding boxes and semantic and instance regions. There were a total of 13 participants in the dataset recordings.
The scene graph itself is annotated using a customized version of CVAT \cite{boris_sekachev_2020_4009388}, which similarly uses a bounding box approach for object detection and annotation.

In total, there are 26 cameras, including 2 ego-view cameras, and 24 $3^{\text{rd}}$ person view cameras to provide a variety of views. Within the $3^{\text{rd}}$
 person views, some video angles comprise a viewpoint looking down from above (ceiling view), and we henceforth count these as part of the $3^{\text{rd}}$ person views. For each long-term video, both participants' ego-view videos are provided, and a subset of the 3rd-person views are provided. Each long-term video contains at least two $3^{\text{rd}}$ person views, with a maximum of 19 $3^{\text{rd}}$ person videos. 


\subsection{Annotation Details}

InteractADL contains (1) high-level activity labels, (2) atomic action class labels, (3) bounding boxes, and (4) spatiotemporal scene-graph labels.  High-level activities encompass many atomic actions, sometimes denoted sub-activity labels in other datasets; for example, a high-level activity of \textit{unloading washing machine} includes the atomic activities \textit{holding basket}, \textit{holding detergent}, etc. We provide temporally localized labels for 50 activity and 334 atomic actions.  InteractADL contains in total 396 activity instances with 5,544 atomic action instances. Bounding boxes for both people and objects are also annotated, totaling 70 classes of bounding boxes (e.g., person, basket, clothes), with 105,697  annotated bounding boxes, of which 3,314  are humans and 102,383  are objects. Finally, we provide densely-annotated scene-graph labels including human-human and human-object interaction labels across 30  different relationship classes. In total, 224,449  relationships are annotated.

Annotated video sequences within InteractADL are quite long, averaging about 10.6  minutes. These long sequence durations allow for complex scene understanding, including action sequences comprised of several ordered atomic actions or longer activities. Atomic actions in InteractADL vary widely in length. While atomic actions are generally short (average of 9.77  seconds), some are as long as 654.9  seconds.  We compare the annotations of InteractADL with existing datasets in Table \ref{tab:dataset_table}.  We note that in our analysis for this work, we focus on atomic action classification from a single view (ego, monitoring, or top-down).   









\section{Method}

\label{sec:method}

Our approach seeks to allow for model-guided separation between semantically similar class names, which are common in fine-grained visual classification datasets. 
Similar class names, which are prevalent when actions happen in constrained environments like in InteractADL, can be challenging for many low-shot classification algorithms designed for VLMs. This is because they often rely on using the given class names from the dataset metadata \cite{ImageCLIP,Coop-zhou2021learning,TIPAdapter,lu2022prompt,ju2022prompting}.  We hypothesize that the default class names for a dataset are likely not the most discriminative for a pre-trained VLM, since they were originally developed for human dataset curation and annotation rather than for model classification.  In a similar way that prompt tuning can be used to improve upon hand-crafted prompts, we show that name tuning improves the given class names for a dataset to make them more discriminative. 

\vspace{-3mm}
\subsection{Method Background and Notation}
\noindent \textbf{Zero-shot Video Classification.} In our work, we focus on the sub-class of VLMs that possess a joint visual-language embedding space (dual encoders), and thus each VLM has a separate visual and text encoder, which we denote as $E^v$ and $E^t$, respectively. We denote $x$ as the input image or video, and $n_i \in \mathbb{R}^{l_i  \times d_\text{token}}$ as the sequence of token embeddings produced by tokenizing the $i^{\text{th}}$ class's name text. In the zero-shot setting, contrastive VLMs perform visual classification by computing the similarity of a visual embedding $E^v(x)$ with text embedding $E^t(n_i)$ for each class name. Usually, a fixed prompt, denoted $q$, is prepended to the class name resulting in text embeddings $E^t([q, n_i])$. The VLM's predicted class probabilities are therefore the softmax of visual-text embedding similarity for each class:
\begin{equation}
        p(y = i \mid x) = \frac{\exp (\frac{1}{\tau} \left<E^t([q, n_i]), E^v(x)\right>)}{\sum_{j=1}^N \exp (\frac{1}{\tau} \left<E^t([q, n_j]), E^v(x)\right>)},
\end{equation}
$\left< \cdot, \cdot \right>$ denotes the VLM-specific similarity function (e.g., cosine similarity or dot-product similarity), and $\tau$ is a VLM-specific temperature parameter that affects the sharpness of the softmax output. 

\noindent \textbf{Prompt Tuning.} Context-Optimization (abbreviated CoOp) was introduced as a method for extending CLIP from zero-shot image classification to few-shot image classification via prompt tuning \cite{Coop-zhou2021learning}.  The method seeks to improve upon prompt engineering approaches by using learnable, fixed-size context vectors instead of using the natural language prompt $q$. 


For CoOp, a context vector sequence $c^\theta \in \mathbb{R}^{l_\text{context} \times d_\text{token}}$ is prepended to every tokenized class name $n_i$, producing text embeddings $E^t([c^\theta, n_i])$. Given a training dataset, the cross-entropy loss is backpropagated through the frozen text encoder to train the context vectors.

\begin{table}[h]
\tablesmall
\centering
\vspace{2mm}
\begin{tabular}{lccc} 

\toprule
  Dataset                                  & Prompt Tuning & Name Tuning &  $\Delta$\\
 \midrule
  MOMA Activities \cite{luo2022moma}       & 95.7   & \textbf{97.9}  & \textcolor{darkgreen}{+2.2}\\
  Kinetics-100   \cite{kay2017kinetics}    & 92.7   & \textbf{94.7}  & \textcolor{darkgreen}{+2.0}\\ 
  MOMA Sub-activities \cite{luo2022moma}   & 76.6   & \textbf{78.2} & \textcolor{darkgreen}{+1.6}\\
\bottomrule 
\end{tabular}
\vspace{3mm}
\caption{\textbf{Comparison of text input tuning methods across few-shot activity recognition benchmarks.} We show show that our proposed method, \textit{name tuning}, consistently outperforms prompt tuning for 5-shot 5-way classification on Kinetics and MOMA-LRG using a CLIP ViT-B32 backbone. Results are averaged across 20 runs with random seeds. }
\label{tab:prompt-comparison}
\end{table}

\subsection{Name Tuning} 
Our method, Name Tuning, seeks to improve the expressiveness of provided class names directly, by training them to better match the content of example videos. 
For each class, we train a class name offset $\epsilon_i^\theta \in \mathbb{R}^{l_\text{class} \times d_\text{token}}$, matching the shape of the class name's token embedding sequence $n_i$ and initialized to zero. Each class name offset is added to the corresponding class name, producing text embeddings $E^t([q, n_i + \epsilon_i^\theta])$.

Like CoOp, we train these parameters using a few-shot training dataset, backpropagating through the frozen text encoder to update the name offset parameters. To avoid overfitting when given only a few training examples per class, we also include an $L_2$ regularization penalty for the name offsets, which constrains the tuned class names to stay near the original class names, retaining the prior information they convey, resulting in the following loss function: 
\vspace{-2mm}
\begin{equation}
J(\theta, \mathcal{D}) = -\sum_{x,y \sim D} y \cdot \log p(y \mid x) + \alpha \frac{1}{2} \sum_{i = 1}^N ||\epsilon_i^\theta||_2^2,
\end{equation}
where $||\cdot||_2$ denotes the $L_2$ norm, and $\alpha$ is a hyperparameter that corresponds to the strength of regularization. We search over $\alpha \in \{0.01, 0.1, 1.0, 10.0\}$ (see Appendix for more details). 

\subsection{CoNa}
Name Tuning is entirely compatible with the existing CoOp method since it tunes separate parts of the input text, and we observe that the combination of the two (abbreviated CoNa) can boost performance on some datasets. CoNa jointly trains a class-shared context $c^\theta \in \mathbb{R}^{l_\text{context} \times d_\text{token}}$ and class-specific name offsets $\epsilon_i^\theta \in \mathbb{R}^{l_\text{class} \times d_\text{token}}$, producing text embeddings $E^t([c^\theta, n_i + \epsilon_i^\theta])$. 

We initialize the elements of the context parameter $c^\theta$ from the normal distribution $\mathcal{N}(\mu,\,\sigma^{2})$, with $\mu=0, \sigma=0.02$ following Zhou~\etal~\cite{Coop-zhou2021learning}, and we initialize the class offset to zero. The $L_2$ regularization penalty is applied only to the name offset parameters.

\vspace{-4mm}
\subsection{Few-shot Learning Paradigm}
To evaluate our methods, we use two formulations of the few-shot learning paradigm.  One which uses a traditional classification dataset with training, validation, and test splits all sharing the same set of classes.  This is the setup used by Radford~\etal~\cite{ImageCLIP} and Zhou~\etal~\cite{Coop-zhou2021learning}.  This formulation contrasts with the meta-learning formulation of few-shot learning, which partitions classification datasets so that training, validation and test sets have entirely separate classes \cite{vinyals2016matching,finn2017model}. Methods in the meta-learning formulation train a conditional classifier on one set of classes so that it can correctly classify unseen classes when provided with only a few labeled examples \cite{finn2017model}. For clear comparisons across datasets and previous works, we use $n$-way $k$-shot to denote $n$ classes in the test set with $k$ support (training) examples per class and explicitly denote when the meta-learning formulation is used.  

\vspace{-4mm}
\subsection{Datasets and Data Splits}
We evaluate Name Tuning and CoNa on five few-shot activity recognition benchmarks: (1) InteractADL, (2) Kinetics \cite{kay2017kinetics}, (3) MOMA-LRG Activities, \cite{luo2022moma}, (4) MOMA-LRG Sub-activities, and (5) UCF-101 \cite{ucf101}.  For Kinetics, we use the classes, videos, and meta-learning splits from Zhu~\etal~\cite{cmn} and use the official few-shot training splits for MOMA-LRG from Luo~\etal~\cite{luo2022moma}. We use standard train/val/test splits for UCF-101 for precise comparison with existing methods and models, and for InteractADL, we split the dataset into a 80/10/20 split and removed activity classes with fewer than 16 videos. 
\vspace{-2mm}
\section{Experiments and Results}
\label{sec:experiments}
Unless denoted otherwise, we use OpenAI's CLIP ViT-B32 checkpoint from \cite{ImageCLIP} for all experiments and results.  Specifically, we encode each frame separately (sampled uniformly) and obtain video-level embeddings by averaging across 10 frames in the video.  For ViFi-CLIP \cite{rasheed2023vificlip}, we use 32 frames and a checkpoint pretrained on Kinetics-400.  For Name Tuning, we use the fixed text prompt: ``a video of \{\}".  We show results on InteractADL in Table \ref{tab:iadl}.  We compare our methods (Name Tuning and CoNa) against strong baselines for few-shot activity recognition with VLMs, namely a linear classifier, VL-prototypes from Luo~\etal~\cite{luo2022moma}, CoOp \cite{Coop-zhou2021learning}, and ViFi-CLIP (few-shot prompt tuning) from Rasheed \etal \cite{rasheed2023vificlip}. Unless stated otherwise, all results from our paper are averaged across 4 random seeds.

\begin{table}[h]
\tablesmall
    \centering
    \begin{tabular}{lrrrr}
        \toprule
        Method & 1 shot & 2 shots & 4 shots & 8 shots \\
        \midrule
        Linear Probe & \textbf{3.95} & 4.94 & 5.41 & 6.71 \\
        VL-Prototype & 3.83 & 5.02 & 4.94 & 5.81 \\
        CoOp & 3.40 & 4.98 & 5.92 & 5.85 \\
        VIFI-CLIP* & 3.36 & 3.88 & 5.50 & 5.97 \\
        \midrule
        Name Tuning & 3.87 & 4.74 & 5.29 & 5.77 \\
        CoNa & 3.87 & \textbf{5.21} & \textbf{6.00} & \textbf{6.95} \\
        \bottomrule
    \end{tabular}
    \vspace{2.5mm}
    \caption{\textbf{56-way few-shot classification results on the InteractADL dataset.} We compare our proposed methods (Name Tuning and CoNa) with existing methods for finetuning VLMs for visual classification.  For fair comparison, we only include text-branch tuning for ViFi-CLIP. 
    Best results in each column are typeset in \textbf{bold}. 
    } 
    \label{tab:iadl}
\end{table}
\vspace{-3mm}
\subsection{Results on Other Activity Recognition Datasets}
In addition to showing results for InteractADL, we demonstrate that Name Tuning and CoNa, without any video pretraining, set a new state-of-the-art for Kinetics (Table \ref{tab:kin100}), and MOMA-LRG (Table \ref{tab:moma}) and provide strong few-shot performance on UCF-101 (Table \ref{tab:ucf101}).  
For the meta-learning splits, we compare against the state-of-the-art meta-learning (vision-only) baselines on the standard 5-shot, 5-way classification task (evaluating unseen classes from the meta-test set).  Our VLM-based method does not make use of the meta-train set, and we choose hyperparameters based on 5-shot 5-way performance on the meta-validation set. Since we do not have a held-out validation set, we do not perform model checkpointing, and simply use the final-epoch model for validation and test results. We report the average test-set accuracy from 20 random seeds. In general, we find that our VLM-based method works well on these meta-learning 
activity recognition datasets and outperforms the vision-only few-shot approaches by a significant margin without requiring a meta-training set. 

\begin{table}[h]
\tablesmall
\centering
\begin{tabular}{lcr}
\toprule
Method & Meta-Training Free & Accuracy \\ 
\midrule
OTAM \cite{otam} & \xmark & 85.8 \\
AMeFu-Net \cite{fu2020depth} & \xmark & 86.8 \\
STRM \cite{thatipelli2021spatio} & \xmark & 86.7 \\
MTFAN \cite{wu2022motion} & \xmark & 87.4 \\
\midrule

  Name Tuning (ours)                    & \checkmark            & \textbf{94.7} \\
  CoNa (ours)                           & \checkmark            & 92.0 \\
\bottomrule
\end{tabular}
\vspace{2.5mm}
\caption{\textbf{5-shot 5-way few-shot classification results for Kinetics.} We compare our proposed methods (Name Tuning and CoNa) with results from previous approaches that rely upon a large meta-training set.  Despite not using the meta-training set, we achieve state-of-the-art results for few-shot activity recognition on Kinetics.  We use the meta-validation set for hyperparameter tuning.}
\label{tab:kin100}
\vspace{-3mm}
\end{table}
\begin{table}[t]
\tablesmall
\centering

\begin{tabular}{lcrr}
\toprule
Method & MT Free & Activity Acc. & Subactivity Acc. \\ 
\midrule
CMN \cite{cmn} & \xmark & 86.3 & 66.6 \\
OTAM \cite{otam} & \xmark & 92.1 & 72.6 \\
VideoCLIP-GVLM \cite{luo2022moma} & \checkmark & 84.8 & 32.7 \\
Frozen-GVLM \cite{luo2022moma} & \checkmark & 92.5 & 26.3 \\
\midrule

  Name Tuning (ours)                    & \checkmark            & \textbf{97.9} & \textbf{78.2} \\
  CoNa (ours)                           & \checkmark            & 95.7 & 76.4 \\
\bottomrule
\end{tabular}
\vspace{2.5mm}
\caption{\textbf{5-shot 5-way few-shot classification results for MOMA-LRG.} We compare our proposed methods (Name Tuning and CoNa) with results from previous meta-learning works. By leveraging pre-trained VLMs, we significantly outperform existing methods and do not require a large meta-training set.  We use the meta-validation set for hyperparameter tuning (MT Free = Meta-Training Free).}
\label{tab:moma}
\end{table}
\begin{table}[t]
\tablesmall
\centering
\begin{tabular}{lrrrr}
\toprule
Method & 1 shot & 2 shots & 4 shots & 8 shots \\
\midrule
Linear Probe & 76.8 & 84.5 & 90.1 & 92.5 \\
VL-Prototype & 79.4 & 86.2 & 91.0 & 92.8 \\
CoOp & 85.3 & 87.1 & 89.8 & 90.8 \\
ViFi-CLIP* & 85.3 & \textbf{88.1} & \textbf{91.5} & 92.6 \\
\midrule
Name Tuning & \textbf{86.3} & 87.5 & 89.1 & 92.7 \\
CoNa & 85.5 & 88.0 & 91.2 & \textbf{93.5} \\
\bottomrule
\end{tabular}
\vspace{2.5mm}
\caption{\textbf{101-way few-shot classification results for UCF-101.} We compare our proposed methods (Name Tuning and CoNa) with existing methods for finetuning VLMs for visual classification. For fair comparison, we only include text-branch tuning for ViFi-CLIP.  We note that ViFi-CLIP requires extensive video pretraining on Kinetics-400 and achieves similar results to CoNa that only uses averaged frame-level CLIP embeddings. We \textbf{bold} the highest accuracy for each number of shots.}
\label{tab:ucf101}
\vspace{-4mm}
\end{table}

\section{Discussion}
\label{sec:discussion}
\paragraph{Comparison of Text Tuning Strategies.}  In Table \ref{tab:prompt-comparison}, we show the average difference in performance between traditional prompt tuning and name tuning across three few-shot activity recognition benchmarks.  We note that the main requisite of CoOp is the existence of reliable and distinguishable class names, whereas Name Tuning requires a good baseline prompt to use to learn class names (for our results, we use ``a video of \{\}"). 

\paragraph{Comparison with Meta-learning Methods.} In our experiments, we find that our VLM-based few-shot learning method outperforms the current state-of-the-art meta-learning approaches on all benchmarked datasets. This is notable since the VLM-based methods do not require a large meta-training set derived from visually similar classes and can instead leverage broad pre-training from internet scale data. 
\paragraph{Performance on InteractADL.}  In Table \ref{tab:iadl}, we show that InteractADL is a very challenging dataset where the benchmarked VLM-based methods perform significantly worse (3-7\%) than other activity recognition datasets (e.g., 76-94\% for UCF-101).  We hypothesize there are several reasons for this gap, including the domain gap between VLM pre-training data and home environment settings (and ego-view data), along with the fine-grained nature and subtle visual distinctions between atomic actions in our dataset.  Regardless, due to the challenging nature of few-shot atomic action classification in home environments (with similar backgrounds) and the rich semantic annotations of InteractADL, we believe that InteractADL provides a challenging benchmark for fine-grained activity recognition at home.

\vspace{-3mm}
\section{Conclusion}
\label{sec:conclusion}

In this paper, we introduced a new dataset, InteractADL, which includes multi-view data of multiple humans performing ADLs at home. To tackle the few-shot activity recognition problem posed by this dataset, we introduce a novel method, Name Tuning, for learning model-specific class names that expands upon existing prompt tuning strategies. Through this method, we also achieve a new state-of-the-art in three different few-shot activity recognition benchmarks (MOMA-Activities, MOMA-Subactivities, and Kinetics).  We demonstrate the challenges for adapting pre-trained VLMs for our fine-grained video dataset, despite their impressive zero-shot performance on relatively high-level datasets.  For transparency, reproducibility, and to accelerate research in this area, we release our code at {\small\url{https://github.com/zanedurante/vlm_benchmark}}.

\section*{Acknowledgments}
Funds to support this AITC study were provided by the Johns Hopkins University AITC under award number P30AG073104. This was also partially supported by Panasonic and the Jaswa Innovator Award. This article solely reflects the opinions and conclusions of its authors and not any entity associated with Johns Hopkins University AITC or Panasonic.
\bibliography{egbib}

\begin{thebibliography}{52}
\providecommand{\natexlab}[1]{#1}
\providecommand{\url}[1]{\texttt{#1}}
\expandafter\ifx\csname urlstyle\endcsname\relax
  \providecommand{\doi}[1]{doi: #1}\else
  \providecommand{\doi}{doi: \begingroup \urlstyle{rm}\Url}\fi

\bibitem[Alayrac et~al.(2022)Alayrac, Donahue, Luc, Miech, Barr, Hasson, Lenc, Mensch, Millican, Reynolds, et~al.]{alayrac2022flamingo}
Jean-Baptiste Alayrac, Jeff Donahue, Pauline Luc, Antoine Miech, Iain Barr, Yana Hasson, Karel Lenc, Arthur Mensch, Katie Millican, Malcolm Reynolds, et~al.
\newblock Flamingo: a visual language model for few-shot learning.
\newblock \emph{arXiv preprint arXiv:2204.14198}, 2022.

\bibitem[Caba~Heilbron et~al.(2015)Caba~Heilbron, Escorcia, Ghanem, and Carlos~Niebles]{caba2015activitynet}
Fabian Caba~Heilbron, Victor Escorcia, Bernard Ghanem, and Juan Carlos~Niebles.
\newblock Activitynet: A large-scale video benchmark for human activity understanding.
\newblock In \emph{Proceedings of the ieee conference on computer vision and pattern recognition}, pages 961--970, 2015.

\bibitem[Cao et~al.(2020)Cao, Ji, Cao, Chang, and Niebles]{otam}
Kaidi Cao, Jingwei Ji, Zhangjie Cao, Chien-Yi Chang, and Juan~Carlos Niebles.
\newblock Few-shot video classification via temporal alignment.
\newblock In \emph{Proceedings of the IEEE/CVF Conference on Computer Vision and Pattern Recognition}, pages 10618--10627, 2020.

\bibitem[Carreira et~al.(2019)Carreira, Noland, Hillier, and Zisserman]{carreira2019short}
Joao Carreira, Eric Noland, Chloe Hillier, and Andrew Zisserman.
\newblock A short note on the kinetics-700 human action dataset.
\newblock \emph{arXiv preprint arXiv:1907.06987}, 2019.

\bibitem[Dai et~al.(2022)Dai, Das, Sharma, Minciullo, Garattoni, Bremond, and Francesca]{dai2022toyota}
Rui Dai, Srijan Das, Saurav Sharma, Luca Minciullo, Lorenzo Garattoni, Francois Bremond, and Gianpiero Francesca.
\newblock Toyota smarthome untrimmed: Real-world untrimmed videos for activity detection.
\newblock \emph{IEEE Transactions on Pattern Analysis and Machine Intelligence}, 2022.

\bibitem[Damen et~al.(2018)Damen, Doughty, Farinella, Fidler, Furnari, Kazakos, Moltisanti, Munro, Perrett, Price, et~al.]{damen2018scaling}
Dima Damen, Hazel Doughty, Giovanni~Maria Farinella, Sanja Fidler, Antonino Furnari, Evangelos Kazakos, Davide Moltisanti, Jonathan Munro, Toby Perrett, Will Price, et~al.
\newblock Scaling egocentric vision: The epic-kitchens dataset.
\newblock In \emph{Proceedings of the European Conference on Computer Vision (ECCV)}, pages 720--736, 2018.

\bibitem[Das et~al.(2019)Das, Dai, Koperski, Minciullo, Garattoni, Bremond, and Francesca]{das2019toyota}
Srijan Das, Rui Dai, Michal Koperski, Luca Minciullo, Lorenzo Garattoni, Francois Bremond, and Gianpiero Francesca.
\newblock Toyota smarthome: Real-world activities of daily living.
\newblock In \emph{Proceedings of the IEEE/CVF International Conference on Computer Vision}, pages 833--842, 2019.

\bibitem[Debes et~al.(2016)Debes, Merentitis, Sukhanov, Niessen, Frangiadakis, and Bauer]{debes2016monitoring}
Christian Debes, Andreas Merentitis, Sergey Sukhanov, Maria Niessen, Nikolaos Frangiadakis, and Alexander Bauer.
\newblock Monitoring activities of daily living in smart homes: Understanding human behavior.
\newblock \emph{IEEE Signal Processing Magazine}, 33\penalty0 (2):\penalty0 81--94, 2016.

\bibitem[Desai et~al.(2004)Desai, Grossberg, and Sheth]{desai2004activities}
Abhilash~K Desai, George~T Grossberg, and Dharmesh~N Sheth.
\newblock Activities of daily living in patients with dementia.
\newblock \emph{CNS drugs}, 18\penalty0 (13):\penalty0 853--875, 2004.

\bibitem[Ehsanpour et~al.(2021)Ehsanpour, Saleh, Savarese, Reid, and Rezatofighi]{ehsanpour2021jrdb}
Mahsa Ehsanpour, Fatemeh Saleh, Silvio Savarese, Ian Reid, and Hamid Rezatofighi.
\newblock Jrdb-act: A large-scale multi-modal dataset for spatio-temporal action, social group and activity detection.
\newblock \emph{arXiv preprint arXiv:2106.08827}, 2021.

\bibitem[Finn et~al.(2017)Finn, Abbeel, and Levine]{finn2017model}
Chelsea Finn, Pieter Abbeel, and Sergey Levine.
\newblock Model-agnostic meta-learning for fast adaptation of deep networks.
\newblock In \emph{International conference on machine learning}, pages 1126--1135. PMLR, 2017.

\bibitem[Fu et~al.(2020)Fu, Zhang, Wang, Fu, and Jiang]{fu2020depth}
Yuqian Fu, Li~Zhang, Junke Wang, Yanwei Fu, and Yu-Gang Jiang.
\newblock Depth guided adaptive meta-fusion network for few-shot video recognition.
\newblock In \emph{Proceedings of the 28th ACM International Conference on Multimedia}, pages 1142--1151, 2020.

\bibitem[Ge et~al.(2022)Ge, Ge, Liu, Wang, Wu, Shan, Qie, and Luo]{ge2022miles}
Yuying Ge, Yixiao Ge, Xihui Liu, Jinpeng Wang, Jianping Wu, Ying Shan, Xiaohu Qie, and Ping Luo.
\newblock Miles: visual bert pre-training with injected language semantics for video-text retrieval.
\newblock In \emph{European Conference on Computer Vision}, pages 691--708. Springer, 2022.

\bibitem[Grauman et~al.(2021)Grauman, Westbury, Byrne, Chavis, Furnari, Girdhar, Hamburger, Jiang, Liu, Liu, et~al.]{grauman2021ego4d}
Kristen Grauman, Andrew Westbury, Eugene Byrne, Zachary Chavis, Antonino Furnari, Rohit Girdhar, Jackson Hamburger, Hao Jiang, Miao Liu, Xingyu Liu, et~al.
\newblock Ego4d: Around the world in 3,000 hours of egocentric video.
\newblock \emph{arXiv preprint arXiv:2110.07058}, 3, 2021.

\bibitem[Gu et~al.(2018)Gu, Sun, Ross, Vondrick, Pantofaru, Li, Vijayanarasimhan, Toderici, Ricco, Sukthankar, et~al.]{gu2018ava}
Chunhui Gu, Chen Sun, David~A Ross, Carl Vondrick, Caroline Pantofaru, Yeqing Li, Sudheendra Vijayanarasimhan, George Toderici, Susanna Ricco, Rahul Sukthankar, et~al.
\newblock Ava: A video dataset of spatio-temporally localized atomic visual actions.
\newblock In \emph{Proceedings of the IEEE Conference on Computer Vision and Pattern Recognition}, pages 6047--6056, 2018.

\bibitem[He et~al.(2021)He, Zhou, Ma, Berg-Kirkpatrick, and Neubig]{he2021towards}
Junxian He, Chunting Zhou, Xuezhe Ma, Taylor Berg-Kirkpatrick, and Graham Neubig.
\newblock Towards a unified view of parameter-efficient transfer learning.
\newblock \emph{arXiv preprint arXiv:2110.04366}, 2021.

\bibitem[Helber et~al.(2019)Helber, Bischke, Dengel, and Borth]{helber2019eurosat}
Patrick Helber, Benjamin Bischke, Andreas Dengel, and Damian Borth.
\newblock Eurosat: A novel dataset and deep learning benchmark for land use and land cover classification.
\newblock \emph{IEEE Journal of Selected Topics in Applied Earth Observations and Remote Sensing}, 2019.

\bibitem[Jang et~al.(2020)Jang, Kim, Park, Jang, Lee, and Kim]{jang2020etri}
Jinhyeok Jang, Dohyung Kim, Cheonshu Park, Minsu Jang, Jaeyeon Lee, and Jaehong Kim.
\newblock Etri-activity3d: A large-scale rgb-d dataset for robots to recognize daily activities of the elderly.
\newblock In \emph{2020 IEEE/RSJ International Conference on Intelligent Robots and Systems (IROS)}, pages 10990--10997. IEEE, 2020.

\bibitem[Ji et~al.(2020)Ji, Krishna, Fei-Fei, and Niebles]{ji2020action}
Jingwei Ji, Ranjay Krishna, Li~Fei-Fei, and Juan~Carlos Niebles.
\newblock Action genome: Actions as compositions of spatio-temporal scene graphs.
\newblock In \emph{Proceedings of the IEEE/CVF Conference on Computer Vision and Pattern Recognition}, pages 10236--10247, 2020.

\bibitem[Ju et~al.(2022)Ju, Han, Zheng, Zhang, and Xie]{ju2022prompting}
Chen Ju, Tengda Han, Kunhao Zheng, Ya~Zhang, and Weidi Xie.
\newblock Prompting visual-language models for efficient video understanding.
\newblock In \emph{European Conference on Computer Vision}, pages 105--124. Springer, 2022.

\bibitem[Katz(1983)]{katz1983assessing}
Sidney Katz.
\newblock Assessing self-maintenance: activities of daily living, mobility, and instrumental activities of daily living.
\newblock \emph{Journal of the American Geriatrics Society}, 1983.

\bibitem[Kay et~al.(2017)Kay, Carreira, Simonyan, Zhang, Hillier, Vijayanarasimhan, Viola, Green, Back, Natsev, Suleyman, and Zisserman]{kay2017kinetics}
Will Kay, Joao Carreira, Karen Simonyan, Brian Zhang, Chloe Hillier, Sudheendra Vijayanarasimhan, Fabio Viola, Tim Green, Trevor Back, Paul Natsev, Mustafa Suleyman, and Andrew Zisserman.
\newblock The kinetics human action video dataset, 2017.

\bibitem[Kong et~al.(2019)Kong, Wu, Deng, Klinkigt, Tong, and Murakami]{kong2019mmact}
Quan Kong, Ziming Wu, Ziwei Deng, Martin Klinkigt, Bin Tong, and Tomokazu Murakami.
\newblock Mmact: A large-scale dataset for cross modal human action understanding.
\newblock In \emph{Proceedings of the IEEE/CVF International Conference on Computer Vision}, pages 8658--8667, 2019.

\bibitem[Kuehne et~al.(2014)Kuehne, Arslan, and Serre]{kuehne2014language}
Hilde Kuehne, Ali Arslan, and Thomas Serre.
\newblock The language of actions: Recovering the syntax and semantics of goal-directed human activities.
\newblock In \emph{Proceedings of the IEEE conference on computer vision and pattern recognition}, pages 780--787, 2014.

\bibitem[Lauretti et~al.(2017)Lauretti, Cordella, Guglielmelli, and Zollo]{lauretti2017learning}
Clemente Lauretti, Francesca Cordella, Eugenio Guglielmelli, and Loredana Zollo.
\newblock Learning by demonstration for planning activities of daily living in rehabilitation and assistive robotics.
\newblock \emph{IEEE Robotics and Automation Letters}, 2\penalty0 (3):\penalty0 1375--1382, 2017.

\bibitem[Li and Liang(2021)]{li2021prefixtuning}
Xiang~Lisa Li and Percy Liang.
\newblock Prefix-tuning: Optimizing continuous prompts for generation.
\newblock In \emph{Proceedings of the 59th Annual Meeting of the Association for Computational Linguistics and the 11th International Joint Conference on Natural Language Processing (Volume 1: Long Papers)}, pages 4582--4597, 2021.

\bibitem[Liu et~al.(2017)Liu, Hu, Li, Song, and Liu]{liu2017pku}
Chunhui Liu, Yueyu Hu, Yanghao Li, Sijie Song, and Jiaying Liu.
\newblock Pku-mmd: A large scale benchmark for continuous multi-modal human action understanding.
\newblock \emph{arXiv preprint arXiv:1703.07475}, 2017.

\bibitem[Liu et~al.(2021)Liu, Yuan, Fu, Jiang, Hayashi, and Neubig]{liu2021promptingmethods}
Pengfei Liu, Weizhe Yuan, Jinlan Fu, Zhengbao Jiang, Hiroaki Hayashi, and Graham Neubig.
\newblock Pre-train, prompt, and predict: A systematic survey of prompting methods in natural language processing.
\newblock \emph{arXiv preprint arXiv:2107.13586}, 2021.

\bibitem[Lu et~al.(2022)Lu, Liu, Zhang, Liu, and Tian]{lu2022prompt}
Yuning Lu, Jianzhuang Liu, Yonggang Zhang, Yajing Liu, and Xinmei Tian.
\newblock Prompt distribution learning.
\newblock In \emph{Proceedings of the IEEE/CVF Conference on Computer Vision and Pattern Recognition}, pages 5206--5215, 2022.

\bibitem[Luo et~al.(2022)Luo, Durante, Li, Xie, Liu, Jin, Huang, Li, Wu, Niebles, et~al.]{luo2022moma}
Zelun Luo, Zane Durante, Linden Li, Wanze Xie, Ruochen Liu, Emily Jin, Zhuoyi Huang, Lun~Yu Li, Jiajun Wu, Juan~Carlos Niebles, et~al.
\newblock Moma-lrg: Language-refined graphs for multi-object multi-actor activity parsing.
\newblock In \emph{Thirty-sixth Conference on Neural Information Processing Systems Datasets and Benchmarks Track}, 2022.

\bibitem[Ni et~al.(2011)Ni, Wang, and Moulin]{ni2011rgbd}
Bingbing Ni, Gang Wang, and Pierre Moulin.
\newblock Rgbd-hudaact: A color-depth video database for human daily activity recognition.
\newblock In \emph{2011 IEEE international conference on computer vision workshops (ICCV workshops)}, pages 1147--1153. IEEE, 2011.

\bibitem[Nilsback and Zisserman(2006)]{oxford_flowers}
Maria-Elena Nilsback and Andrew Zisserman.
\newblock A visual vocabulary for flower classification.
\newblock In \emph{IEEE Conference on Computer Vision and Pattern Recognition}, volume~2, pages 1447--1454, 2006.

\bibitem[Parkhi et~al.(2012)Parkhi, Vedaldi, Zisserman, and Jawahar]{oxford_pets}
Omkar~M. Parkhi, Andrea Vedaldi, Andrew Zisserman, and C.~V. Jawahar.
\newblock Cats and dogs.
\newblock In \emph{IEEE Conference on Computer Vision and Pattern Recognition}, 2012.

\bibitem[Radford et~al.(2021)Radford, Kim, Hallacy, Ramesh, Goh, Agarwal, Sastry, Askell, Mishkin, Clark, Krueger, and Sutskever]{ImageCLIP}
Alec Radford, Jong~Wook Kim, Chris Hallacy, Aditya Ramesh, Gabriel Goh, Sandhini Agarwal, Girish Sastry, Amanda Askell, Pamela Mishkin, Jack Clark, Gretchen Krueger, and Ilya Sutskever.
\newblock Learning transferable visual models from natural language supervision.
\newblock In \emph{ICML}, 2021.

\bibitem[Rai et~al.(2021)Rai, Chen, Ji, Desai, Kozuka, Ishizaka, Adeli, and Niebles]{rai2021home}
Nishant Rai, Haofeng Chen, Jingwei Ji, Rishi Desai, Kazuki Kozuka, Shun Ishizaka, Ehsan Adeli, and Juan~Carlos Niebles.
\newblock Home action genome: Cooperative compositional action understanding.
\newblock In \emph{Proceedings of the IEEE/CVF Conference on Computer Vision and Pattern Recognition}, pages 11184--11193, 2021.

\bibitem[Rasheed et~al.(2023)Rasheed, Khattak, Maaz, Khan, and Khan]{rasheed2023vificlip}
Hanoona Rasheed, Muhammad~Uzair Khattak, Muhammad Maaz, Salman Khan, and Fahad~Shahbaz Khan.
\newblock Fine-tuned clip models are efficient video learners, 2023.

\bibitem[Sekachev et~al.(2020)Sekachev, Manovich, Zhiltsov, Zhavoronkov, Kalinin, Hoff, TOsmanov, Kruchinin, Zankevich, DmitriySidnev, Markelov, Johannes222, Chenuet, a~andre, telenachos, Melnikov, Kim, Ilouz, Glazov, Priya4607, Tehrani, Jeong, Skubriev, Yonekura, vugia truong, zliang7, lizhming, and Truong]{boris_sekachev_2020_4009388}
Boris Sekachev, Nikita Manovich, Maxim Zhiltsov, Andrey Zhavoronkov, Dmitry Kalinin, Ben Hoff, TOsmanov, Dmitry Kruchinin, Artyom Zankevich, DmitriySidnev, Maksim Markelov, Johannes222, Mathis Chenuet, a~andre, telenachos, Aleksandr Melnikov, Jijoong Kim, Liron Ilouz, Nikita Glazov, Priya4607, Rush Tehrani, Seungwon Jeong, Vladimir Skubriev, Sebastian Yonekura, vugia truong, zliang7, lizhming, and Tritin Truong.
\newblock opencv/cvat: v1.1.0, August 2020.
\newblock URL \url{https://doi.org/10.5281/zenodo.4009388}.

\bibitem[Sigurdsson et~al.(2016)Sigurdsson, Varol, Wang, Farhadi, Laptev, and Gupta]{sigurdsson2016hollywood}
Gunnar~A Sigurdsson, G{\"u}l Varol, Xiaolong Wang, Ali Farhadi, Ivan Laptev, and Abhinav Gupta.
\newblock Hollywood in homes: Crowdsourcing data collection for activity understanding.
\newblock In \emph{Computer Vision--ECCV 2016: 14th European Conference, Amsterdam, The Netherlands, October 11--14, 2016, Proceedings, Part I 14}, pages 510--526. Springer, 2016.

\bibitem[Sigurdsson et~al.(2018)Sigurdsson, Gupta, Schmid, Farhadi, and Alahari]{sigurdsson2018charades}
Gunnar~A Sigurdsson, Abhinav Gupta, Cordelia Schmid, Ali Farhadi, and Karteek Alahari.
\newblock Charades-ego: A large-scale dataset of paired third and first person videos.
\newblock \emph{arXiv preprint arXiv:1804.09626}, 2018.

\bibitem[Soomro et~al.(2012)Soomro, Zamir, and Shah]{ucf101}
Khurram Soomro, Amir~Roshan Zamir, and Mubarak Shah.
\newblock Ucf101: A dataset of 101 human actions classes from videos in the wild.
\newblock \emph{arXiv preprint arXiv:1212.0402}, 2012.

\bibitem[Thatipelli et~al.(2022)Thatipelli, Narayan, Khan, Anwer, Khan, and Ghanem]{thatipelli2021spatio}
Anirudh Thatipelli, Sanath Narayan, Salman Khan, Rao~Muhammad Anwer, Fahad~Shahbaz Khan, and Bernard Ghanem.
\newblock Spatio-temporal relation modeling for few-shot action recognition.
\newblock In \emph{CVPR}, 2022.

\bibitem[Vaquette et~al.(2017)Vaquette, Orcesi, Lucat, and Achard]{vaquette2017daily}
Geoffrey Vaquette, Astrid Orcesi, Laurent Lucat, and Catherine Achard.
\newblock The daily home life activity dataset: a high semantic activity dataset for online recognition.
\newblock In \emph{2017 12th IEEE International Conference on Automatic Face \& Gesture Recognition (FG 2017)}, pages 497--504. IEEE, 2017.

\bibitem[Vinyals et~al.(2016)Vinyals, Blundell, Lillicrap, Wierstra, et~al.]{vinyals2016matching}
Oriol Vinyals, Charles Blundell, Timothy Lillicrap, Daan Wierstra, et~al.
\newblock Matching networks for one shot learning.
\newblock \emph{Advances in neural information processing systems}, 29, 2016.

\bibitem[Wang et~al.(2012)Wang, Liu, Wu, and Yuan]{wang2012mining}
Jiang Wang, Zicheng Liu, Ying Wu, and Junsong Yuan.
\newblock Mining actionlet ensemble for action recognition with depth cameras.
\newblock In \emph{2012 IEEE conference on computer vision and pattern recognition}, pages 1290--1297. IEEE, 2012.

\bibitem[Wu et~al.(2022)Wu, Zhang, Zhang, Wu, and Zhang]{wu2022motion}
Jiamin Wu, Tianzhu Zhang, Zhe Zhang, Feng Wu, and Yongdong Zhang.
\newblock Motion-modulated temporal fragment alignment network for few-shot action recognition.
\newblock In \emph{Proceedings of the IEEE/CVF Conference on Computer Vision and Pattern Recognition}, pages 9151--9160, 2022.

\bibitem[Xu et~al.(2021)Xu, Ghosh, Huang, Okhonko, Aghajanyan, Metze, Zettlemoyer, and Feichtenhofer]{VideoCLIP}
Hu~Xu, Gargi Ghosh, Po-Yao Huang, Dmytro Okhonko, Armen Aghajanyan, Florian Metze, Luke Zettlemoyer, and Christoph Feichtenhofer.
\newblock Videoclip: Contrastive pre-training for zero-shot video-text understanding.
\newblock In \emph{Proceedings of the 2021 Conference on Empirical Methods in Natural Language Processing}, pages 6787--6800, 2021.

\bibitem[Yu et~al.(2022)Yu, Wang, Vasudevan, Yeung, Seyedhosseini, and Wu]{yu2022coca}
Jiahui Yu, Zirui Wang, Vijay Vasudevan, Legg Yeung, Mojtaba Seyedhosseini, and Yonghui Wu.
\newblock Coca: Contrastive captioners are image-text foundation models.
\newblock \emph{arXiv preprint arXiv:2205.01917}, 2022.

\bibitem[Zhang et~al.(2020)Zhang, Gould, and Ben-Shabat]{zhang2020vidat}
Jiahao Zhang, Stephen Gould, and Itzik Ben-Shabat.
\newblock Vidat---{ANU} {CVML} video annotation tool.
\newblock \url{https://github.com/anucvml/vidat}, 2020.

\bibitem[Zhang et~al.(2021)Zhang, Fang, Zhang, Gao, Li, Dai, Qiao, and Li]{TIPAdapter}
Renrui Zhang, Rongyao Fang, Wei Zhang, Peng Gao, Kunchang Li, Jifeng Dai, Yu~Qiao, and Hongsheng Li.
\newblock Tip-adapter: Training-free clip-adapter for better vision-language modeling.
\newblock \emph{CoRR}, abs/2111.03930, 2021.
\newblock URL \url{https://arxiv.org/abs/2111.03930}.

\bibitem[Zhou et~al.(2021)Zhou, Yang, Loy, and Liu]{Coop-zhou2021learning}
Kaiyang Zhou, Jingkang Yang, Chen~Change Loy, and Ziwei Liu.
\newblock Learning to prompt for vision-language models.
\newblock \emph{arXiv preprint arXiv:2109.01134}, 2021.

\bibitem[Zhou et~al.(2022)Zhou, Yang, Loy, and Liu]{zhou2022conditional}
Kaiyang Zhou, Jingkang Yang, Chen~Change Loy, and Ziwei Liu.
\newblock Conditional prompt learning for vision-language models.
\newblock In \emph{Proceedings of the IEEE/CVF Conference on Computer Vision and Pattern Recognition}, pages 16816--16825, 2022.

\bibitem[Zhu and Yang(2018)]{cmn}
Linchao Zhu and Yi~Yang.
\newblock Compound memory networks for few-shot video classification.
\newblock In \emph{Proceedings of the European Conference on Computer Vision (ECCV)}, pages 751--766, 2018.

\end{thebibliography}
\clearpage
\part*{Appendix}
\appendix
\begin{figure}[h]
    \centering
    {\includegraphics[width=0.65\linewidth]{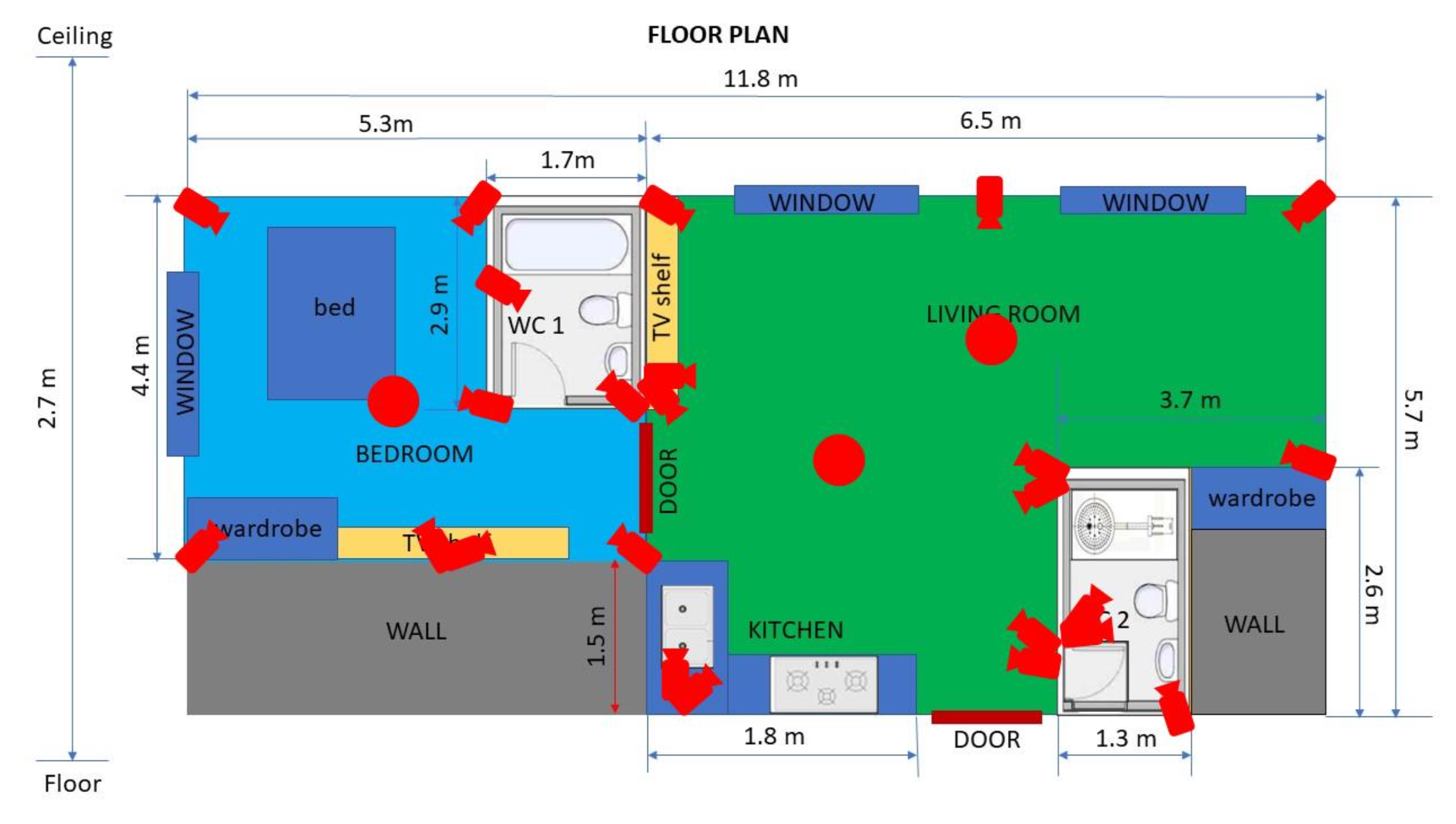}
    }
    \caption{We detail the floorplan of the home environment used to collect data for InteractADL.  InteractADL is recorded in a real home environment with sensors (represented in red) placed throughout the many rooms to provide a subset of views for each long-term video, including ego-view, $3^\text{rd}$ person, and top-down. Ego-view is provided for all videos, and at least two $3^\text{rd}$ person views (including ceiling views) are provided for each video.}
    \label{fig:floorplan}
\end{figure} 

\section{InteractADL Details}

\begin{figure}[h]
    \centering
    {\includegraphics[width=0.80\linewidth]{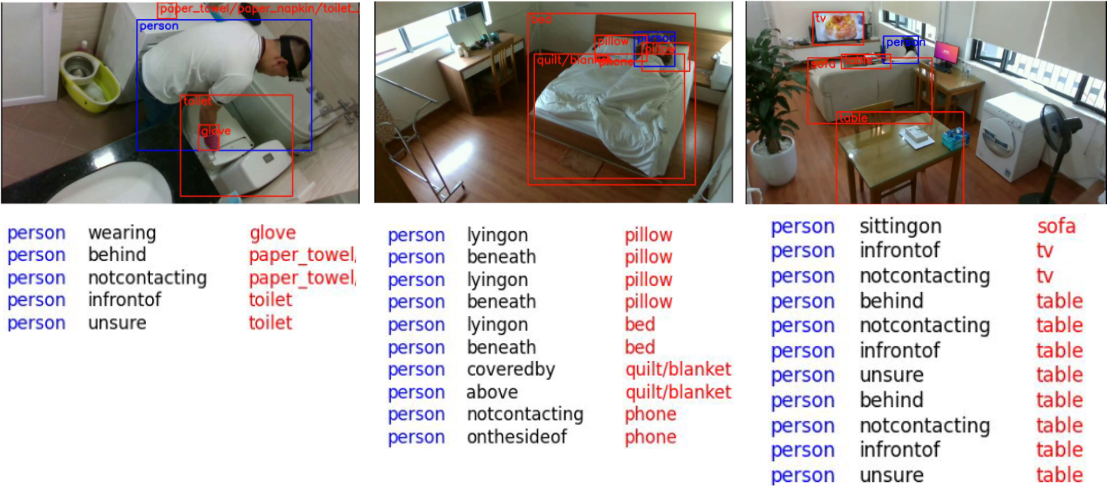}
    }
    \vspace{3mm}
    \caption{We show three  example scene graph annotations from InteractADL corresponding to three separate rooms in the home. Our annotations cover the primary household objects and actors in the scene and  interactions relevant for ADLs.}
    \label{fig:sample_frames}
\end{figure}

\textbf{Floorplan} We show a detailed floorplan of a real home that was used for collecting data for InteractADL in Figure \ref{fig:floorplan}.
\textbf{Dataset Samples} We show three example frames from our labeled dataset in Figure \ref{fig:sample_frames}.

\section{Name Tuning Ablation and Analysis}
\label{sec:ablation}
A majority of our ablations used either the Kinetics or MOMA-LRG datasets.  We chose these datasets since (1) Kinetics is a widely used benchmark for video understanding and activity recognition, and (2) MOMA-LRG was a dataset specifically constructed for challenging VLMs to require more fine-grained understanding of video scenes.

\begin{figure}[t]
    \centering
    \label{fig:model}
    \includegraphics[width=0.8\linewidth]{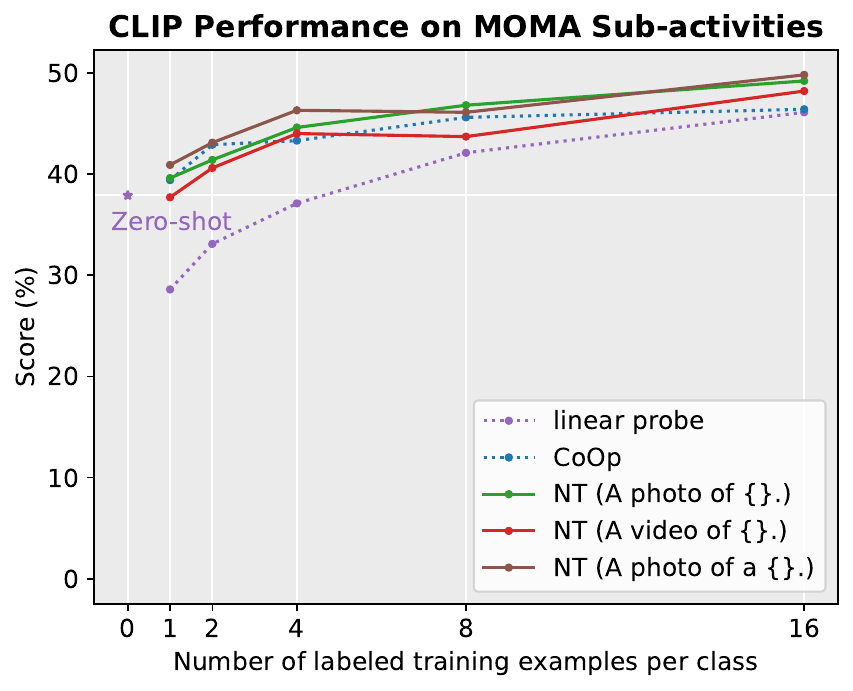}
    \caption{Name Tuning performance on the MOMA Sub-activities dataset using various prompts and CLIP as a pre-trained backbone.  We label the prompt besides Name Tuning (NT) in the legend and show linear probe and CoOp as baselines.}
    \label{fig:prompt-effect}
\end{figure}

\begin{figure}[t]
    \centering
    \includegraphics[width=\linewidth]{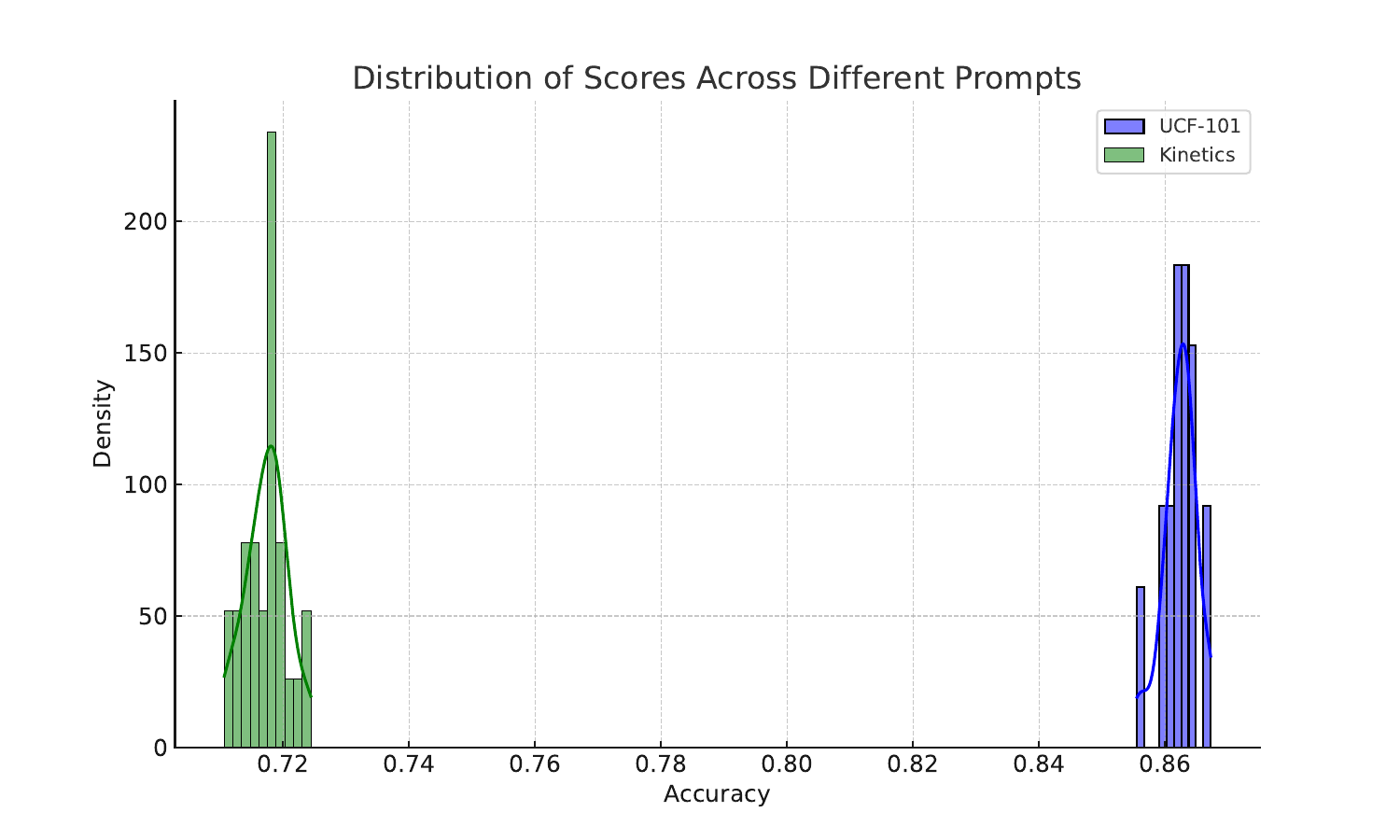}
    \caption{Name Tuning performance for 5-shot classification on Kinetics and UCF101. We use the 28 prompt templates from Radford~\etal~\cite{ImageCLIP} originally used for the UCF-101 dataset to calculate a distribution of Name Tuning scores across various prompts.  In general, prompt choice can provide small increases in performance (up to 2\% absolute).}
    \label{fig:prompt-distribution}

\end{figure}

\textbf{Effect of Prompt Choice.} To help better understand the effect of prompt choice on the performance of Name Tuning, we show the performance of three different prompts on MOMA Sub-activities in Figure \ref{fig:prompt-effect}.  We also show how varying the prompt for Name Tuning can effect 5-shot all-way performance on UCF-101 and Kinetics in Figure \ref{fig:prompt-distribution}. For this analysis, to better observe the effects of prompt choice and disentangle other hyperparameter choices, we fix the learning rate to $10^{-4}$ and the name regularization parameter to 1 and thus do not perform any hyperparameter search, resulting in slightly worse performance than what we report in the main paper.  For Figure \ref{fig:prompt-distribution}, we use the 28 prompts from \cite{ImageCLIP} that were originally used to obtain zero-shot performance for the UCF-101 dataset.  We show results for each prompt averaged across 20 runs.  To summarize our findings: good prompt choice can result in small improvements in performance (up to 2\% absolute). For the exact prompts used, see the prompt templates described in \url{https://github.com/openai/CLIP/blob/main/data/prompts.md#ucf101}.

\begin{figure}[t]
    \centering
    \label{fig:model}
    \includegraphics[width=0.8\linewidth]{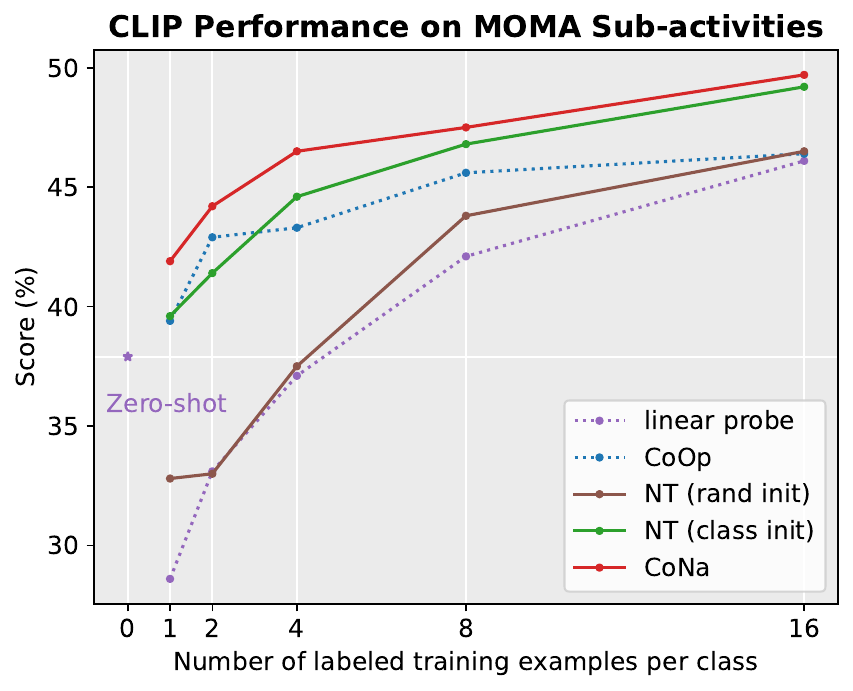}
    \caption{Performance on the MOMA Sub-activities dataset for various VLM fine-tuning methods using CLIP as a pre-trained backbone.  For Name Tuning (NT), we use "A video of \{\}". We show that using randomly initialized (rand init) class names under-performs using the default dataset class names (class init) for Name Tuning.}
    \label{fig:random-init}
\end{figure}

\textbf{Randomly Initialized Class Names.} We also explored how important the initialization of class names is for Name Tuning performance.  Thus, we test the performance on MOMA Sub-activities with class names randomly initialized to $\mathcal{N}(0, 0.02)$, following the initialization procedure in CoOp \cite{Coop-zhou2021learning}. Since we do not use a semantically meaningful initialization for class names, we do not use the regularization term to penalize large learned offset vectors, thus allowing greater exploration of the input vector space.  We show that random initialization under-performs initializing input vectors to the default dataset class names and report results for 1, 2, 4, 8, and 16 shots in Figure \ref{fig:random-init}.

\textbf{Results on Image Classification Datasets}
\label{sec:image-level}
We verify that Name Tuning can also perform well on fine-grained image classification datasets when using CLIP as our backbone.  We select a subset of the datasets from \cite{Coop-zhou2021learning} that are and challenging for zero-shot CLIP and visually fine-grained. Namely, we use Oxford Flowers \cite{oxford_flowers}, Oxford Pets \cite{oxford_pets}, and EuroSAT \cite{helber2019eurosat}.  Generally, Name Tuning performs similarly to CoOp for the datasets evaluated.  We show results in Table \ref{tab:image_level}.

\begin{table}[t!]

 \centering
  \begin{tabular}{llrrrrr}
   \toprule
    Dataset &
    Method &
    1-shot &
    2-shot &
    4-shot &
    8-shot &
    16-shot \\
    
    \midrule

    Oxford Flowers &
    CoOp \cite{Coop-zhou2021learning} &
    74.9 &
    80.4 &
    86.8 &
    91.4 &
    93.6 \\

    &
    Name Tuning &
    76.2 &
    83.4 &
    89.0 &
    92.4 &
    95.0 \\

    \midrule

    Oxford Pets &
    CoOp \cite{Coop-zhou2021learning} &
    87.1 &
    86.4 &
    87.8 &
    89.6 &
    90.1 \\

    &
    Name Tuning &
    82.7 &
    84.8 &
    86.5 &
    87.5 &
    89.0 \\

    \midrule

    EuroSAT &
    CoOp \cite{Coop-zhou2021learning} &
    54.4 &
    59.8 &
    73.8 &
    80.6 &
    83.9 \\

    &
    Name Tuning &
    54.3 &
    66.6 &
    72.6 &
    82.2 &
    85.1 \\

    \bottomrule

    \end{tabular}
    \vspace{2.5mm}

\caption{\textbf{Name Tuning and CoOp results on few-shot image classification.} Both methods are evaluated using CLIP ViT-B32 as the pretrained VLM. Generally, Name Tuning performs similar to CoOp for fine-grained, few-shot image classification datasets with semantically similar classes.
}
\label{tab:image_level}
\end{table}
\subsection{Name Tuning Comparison to CoOp-CSC}

CoOp-CSC is a version of the CoOp algorithm that trains class-specific context (CSC) embedding for each visual category. The authors find that it generally performs worse than the shared-context version of CoOp, though they note that it has potential in fine-grained environments \cite{Coop-zhou2021learning}. In Table \ref{tab:csc-comparison}, we compare CoOp-CSC against both CoOp and our algorithm CoNa for the fine-grained dataset  MOMA Sub-activities. Our results show that shared-context with name-tuning (CoNa, ours) outperforms CoOp both with shared-context and class-specific-context training across most task settings in both datasets. Though the performance difference is large for low-shot settings, the algorithms tend to have closer performance in the 8 and 16-shot settings. 
Although both CoNa and CoOp-CSC have the ability to learn class-specific variables, CoOp-CSC must learn them from scratch whereas CoNa is given a helpful starting point by leveraging the dataset class names with an offset initialized to zero. CoOp-CSC also does not share information between classes, and thus can only learn class-specific variables, whereas CoNa uses shared context vectors.  We hypothesize these factors may contribute to why CoNa performs better than CoOp-CSC in the few-shot setting. 

\begin{table*}[t!]
{
 \centering
 \resizebox{1.0\textwidth}{!}{%
  \begin{tabular}{lccccc}
   \toprule

    Method &
   1-shot &
   2-shot &
   4-shot &
   8-shot &
   16-shot \\
   
   \midrule
   \midrule

    CoOp  &
    39.4, \textbf{37.2}, 23.1 & 
    42.9, 30.3, 26.1 &
    43.3, 38.3, 31.6 &
    45.6, 40.4, 38.6 &
    46.4, 43.6, 37.7 \\
    
    CoOp-CSC  &
    31.5, 33.3, 26.5 &
    34.9, 38.0, 31.1 &
    37.4, 38.7, 33.8 &
    40.5, 42.8, \textbf{38.7} &
    44.7, 45.7, 40.9 \\
    
    Name Tuning &  
    39.6, 36.5, \textbf{29.6} &
    41.4, \textbf{38.8}, \textbf{33.6} &
    44.6, 40.5, \textbf{37.6} &
    46.8, \textbf{43.3}, 38.0 &
    49.2, \textbf{46.3}, \textbf{41.7} \\
    CoNa  &
    \textbf{41.9}, 37.1, 29.2 &
    \textbf{44.2}, 38.5, 32.6 &
    \textbf{46.5}, \textbf{40.9}, 36.9 &
    \textbf{47.5}, 42.2, \textbf{38.7} &
    \textbf{49.7}, 45.9, 40.9 \\
    \bottomrule
  \end{tabular}
  }
  \vspace{2mm}
   \caption{\textbf{Comparison vs CoOp (CSC) on the MOMA-Subactivities activity recognition dataset.}  We show results using CLIP, MILES, and VideoCLIP by denoting each entry with three values representing the performance of the three VLMs, respectively.  We \textbf{bold} the highest accuracy for each VLM, and for each number of shots.}
  \label{tab:csc-comparison}
}
\end{table*}
\section{Results Across VLMs}
\begin{table*}[t!]

 \centering
  \resizebox{1.0\textwidth}{!}{%

  \begin{tabular}{lcccccc}
   \toprule
    Method &
   0-shot &
   1-shot &
   2-shot &
   4-shot &
   8-shot &
   16-shot \\
   
   \midrule

      Prompt Engineering  &
    81.3, 70.7, 49.5 &
    81.3, 70.7, 49.5 & 
    81.3, 70.7, 49.5 &
    81.3, 70.7, 49.5 &
    81.3, 70.7, 49.5 &
    81.3, 70.7, 49.5 \\
   
   VL-Prototype  &
   ---  &


   85.8, 75.7, 60.7 &
   87.5, \textbf{80.3}, 63.4 &
   85.5, 84.2, 65.4 &
   88.4, 86.4, 67.9 &
   90.5, 88.5, 69.7 \\
   
   Linear Probe  &
    --- &
    67.2, 70.6, 54.6 &
    80.6, 77.9, 61.7 &
    86.0, 81.7, 70.5 &
    90.8, \textbf{87.8}, 77.1 &
    93.0, \textbf{89.6}, \textbf{79.2} \\

   CoOp  &
    --- &
    88.7, 73.7, 59.0 &
    90.9, 78.8, 61.4 &
    91.7, 82.1, 66.8 &
    93.6, 84.3, 73.1 &
    95.5, 86.6, 77.7 \\
    \midrule

    Name Tuning (ours) &
--- & 
\textbf{88.9}, \textbf{78.4}, \textbf{61.9} &
\textbf{91.7}, 79.8, \textbf{67.0} &
91.0, \textbf{84.6}, \textbf{73.3} &
93.7, 85.9, \textbf{74.9} &
\textbf{95.8}, 87.8, 77.7 \\
   
    CoNa (ours)  &
    --- &
    87.9, 75.1, 58.8 &
    91.3, 78.9, 63.9 &
    \textbf{92.3}, 82.1, 70.7 &    
    \textbf{94.0}, 82.5, 71.8 &
    94.7, 86.5, 78.8\\
    

   \bottomrule
  \end{tabular}%
  }
  \vspace{3mm}
   \caption{\textbf{Low-shot classification results on the MOMA-LRG Activities dataset.}  We show results using CLIP, MILES, and VideoCLIP by denoting each entry with three values representing the performance of the three VLMs, respectively.  We \textbf{bold} the highest accuracy for each VLM and for each number of shots.}
     \label{tab:complete-moma-act}

\end{table*}
\begin{table*}[t!]

 \centering
 \resizebox{1.0\textwidth}{!}{%
  \begin{tabular}{lcccccc}
   \toprule
    Method &
   0-shot &
   1-shot &
   2-shot &
   4-shot &
   8-shot &
   16-shot \\
   
   \midrule

   Prompt Engineering  &
    37.9, 32.0, 23.8 &
    37.9, 32.0, 23.8 & 
    37.9, 32.0, 23.8 &
    37.9, 32.0, 23.8 &
    37.9, 32.0, 23.8 &
    37.9, 32.0, 23.8 \\
   
   VL-Prototype  &
   ---- &


   40.0, \textbf{37.8}, 27.9 &
   40.8, \textbf{40.0}, 30.6 &
   41.5, \textbf{41.6}, 32.7 &
   41.6, 42.7, 34.3 &
   42.3, 44.3, 36.7 \\
   
   Linear Probe  &
    --- &
    28.6, 31.0, 26.2 &
    33.1, 35.2, 31.0 &
    37.1, 38.9, 36.1 &
    42.1, 42.0, 38.6 &
    46.1, 44.8, \textbf{42.5} \\
   
   CoOp  &
    --- &
    39.4, 37.2, 23.1 & 
    42.9, 30.3, 26.1 &
    43.3, 38.3, 31.6 &
    45.6, 40.4, 38.6 &
    46.4, 43.6, 37.7 \\
    \midrule

        Name Tuning (ours) &
--- & 
39.6, 36.5, \textbf{29.6} &
41.4, 38.8, \textbf{33.6} &
44.6, 40.5, \textbf{37.6} &
46.8, \textbf{43.3}, 38.0 &
49.2, \textbf{46.3}, 41.7 \\
    
   
    CoNa (ours)  &
    --- &
    \textbf{41.9}, 37.1, 29.2 &
    \textbf{44.2}, 38.5, 32.6 &
    \textbf{46.5}, 40.9, 36.9 &
    \textbf{47.5}, 42.2, \textbf{38.7} &
    \textbf{49.7}, 45.9, 40.9 \\


   \bottomrule
  \end{tabular}
}
\vspace{3mm}
   \caption{\textbf{Low-shot classification results on the MOMA-LRG Sub-activities dataset.} We show results using CLIP, MILES, and VideoCLIP by denoting each entry with three values representing the performance of the three VLMs, respectively.  We \textbf{bold} the highest accuracy for each VLM and for each number of shots. }
     \label{tab:complete-moma-sact}

\end{table*}
\begin{table*}[t!]
 
\centering
 \resizebox{1.0\textwidth}{!}{%
  \begin{tabular}{lcccccc}
   \toprule
    Method &
   0-shot &
   1-shot &
   2-shot &
   4-shot &
   8-shot &
   16-shot \\
   
   \midrule

    Prompt Engineering  &
    72.4, 54.6, 37.6 &
    72.4, 54.6, 37.6  & 
    72.4, 54.6, 37.6  &
    72.4, 54.6, 37.6  &
    72.4, 54.6, 37.6  &
    72.4, 54.6, 37.6  \\
   
   VL-Prototype  &
   --- &


   72.2, 59.3, 41.2 &
   72.3, \textbf{61.9}, 44.5 &
   72.2, \textbf{64.2}, 46.5 &
   71.9, 66.6, 48.5 &
   70.9, 68.2, 50.5 \\
   
   Linear Probe  &
    --- &
   37.0, 45.9, 28.7 &
   49.8, 55.2, 38.2 &
   58.3, 61.9, 48.3 &
   65.9, 66.8, \textbf{54.9} &
   71.1, \textbf{70.6}, 57.8 \\

   CoOp  &
    --- &
    70.0, 59.1, 40.1 &
    72.1, 60.7, 41.4 &
    73.0, 63.4, 47.2 &
    75.3, 66.6, 51.5 &
    76.1, 68.2, 56.0 \\
    \midrule
        Name Tuning (ours) &
--- & 
\textbf{73.2}, 58.3, \textbf{42.3} &
\textbf{73.8}, 60.7, \textbf{45.6} &
\textbf{74.7}, 63.7, \textbf{49.0} &
\textbf{76.0}, 66.5, 53.3 &
77.7, 69.2, 57.4 \\
   
    CoNa (ours)  &
    --- &
    69.9, 59.0, 41.3 &
    71.7, 61.5, 44.2 &
    73.4, 63.7, 47.0 &
    \textbf{76.0}, 66.8, 52.1 &
    \textbf{77.8}, 68.7, 55.8\\
   
   
   \bottomrule
  \end{tabular}
  }
  \vspace{3mm}
  \caption{\textbf{Low-shot classification results on the Kinetics-100 dataset.} We show results using CLIP, MILES, and VideoCLIP by denoting each entry with three values representing the performance of the three VLMs, respectively.  We \textbf{bold} the highest accuracy for each VLM and for each number of shots.}
  \label{tab:complete-kinetics}
\end{table*}

We show results for 3 dual encoders VLMs in our experiments. Specifically, we try CLIP \cite{ImageCLIP}, VideoCLIP \cite{VideoCLIP}, and MILES \cite{ge2022miles}.  Results can be seen in Tables  \ref{tab:complete-moma-act}, \ref{tab:complete-moma-sact}, and \ref{tab:complete-kinetics}, where our methods (Name Tuning and CoNa) generalize to VLMs beyond CLIP and perform well on all benchmarked datasets. CLIP outperforms MILES and VideoCLIP as a pre-trained VLM across all methods and datasets tested, which we found to be particularly notable since CLIP can only be used as a frame-level encoder.  We also show results on the 5-shot, 5-way task for each VLM backbone in Tables \ref{tab:complete-meta-learning-kinetics}, \ref{tab:complete-meta-learning-moma-act}, and \ref{tab:complete-meta-learning-moma-sact}.  Similarly, CLIP outperforms the other VLMs and Name Tuning outperforms all other benchmarked methods.

%
 %
 
  %
 
%

\begin{table}[h
]
\tablesmall
\centering

\begin{tabular}{lcc}
\toprule
  Method              & Meta-training Free? & Accuracy        \\
 \midrule
  OTAM   \cite{otam}                            &       \xmark     &   85.8                \\ 
  AMeFu-Net \cite{fu2020depth}                  &       \xmark     &   86.8                       \\
  \midrule
  VL-Prototype                          & \checkmark            & 92.9, 93.5, 85.0 \\
  Linear Probe                          & \checkmark            & 89.2, 92.7, 86.8 \\
  CoOp \cite{Coop-zhou2021learning}     & \checkmark            & 92.7, 91.7, 86.7 \\
  Name Tuning (ours)                          & \checkmark            & \textbf{94.7}, 90.8, 85.8 \\
  CoNa (ours)                           & \checkmark            & 92.0, 92.0, 86.5 \\

\bottomrule
\end{tabular}
\vspace{3mm}
\caption{\textbf{Comparison vs state-of-the-art meta-learning algorithms for the Kinetics dataset.}  We show performance using standard 5-shot 5-way accuracy on the meta-test set. We note that the VLM-based few-shot learning approaches do not require a meta-training set.  For the VLM-based approaches, we show results using CLIP, MILES, and VideoCLIP by denoting each entry with three values representing the performance of the three VLMs, respectively. We \textbf{bold} the single best accuracy in the table.}
\label{tab:complete-meta-learning-kinetics}
\end{table}
%
 %
 
  %
 
%

\begin{table}[h]
\tablesmall
\centering

\begin{tabular}{lcc}
\toprule
  Method              & Meta-training Free? & Accuracy        \\
 \midrule
  CMN   \cite{cmn}                              &        \xmark & 86.3            \\ 
  OTAM \cite{otam}                              &        \xmark & 92.07                      \\
  \midrule
  VL-Prototype                          & \checkmark            & 86.5, 95.9, 92.3 \\
  Linear Probe                          & \checkmark            & 87.4, 96.1, 75.9 \\
  CoOp \cite{Coop-zhou2021learning}     & \checkmark            & 88.6, 95.7, 92.8 \\
  Name Tuning (ours)                          & \checkmark            & \textbf{97.9}, 95.6, 92.7 \\
  CoNa (ours)                           & \checkmark            & 90.0, 95.7, 92.9 \\

\bottomrule
\end{tabular}
\vspace{3mm}
\caption{\textbf{Comparison vs state-of-the-art meta-learning algorithms for the MOMA Activities dataset.}  We show performance using standard 5-shot 5-way accuracy on the meta-test set. We note that the VLM-based few-shot learning approaches do not require a meta-training set.  For the VLM-based approaches, we show results using CLIP, MILES, and VideoCLIP by denoting each entry with three values representing the performance of the three VLMs, respectively. We \textbf{bold} the single best accuracy in the table.}
\label{tab:complete-meta-learning-moma-act}
\end{table}
%
 %
 
  %
 
%

\begin{table}[h]
\tablesmall
\centering
\begin{tabular}{lcc}
\toprule
  Method              & Meta-training Free? & Accuracy        \\
 \midrule
  CMN   \cite{cmn}                              &       \xmark & 66.6                \\ 
  OTAM \cite{otam}                              &         \xmark & 72.6                       \\
  \midrule
  VL-Prototype                          & \checkmark            & 71.2, 75.7, 75.6 \\
  Linear Probe                          & \checkmark            & 71.3, 75.1, 74.5 \\
  CoOp \cite{Coop-zhou2021learning}     & \checkmark            & 74.0, 72.8, 76.6 \\
    Name Tuning (ours)                          & \checkmark            & 71.0, 74.0, \textbf{78.2} \\

  CoNa (ours)                           & \checkmark            & 71.3, 73.6, 76.4 \\

\bottomrule
\end{tabular}

\vspace{3mm}
\caption{\textbf{Comparison vs state-of-the-art meta-learning algorithms for the MOMA Sub-activities dataset.}  We show performance using standard 5-shot 5-way accuracy on the meta-test set. We note that the VLM-based few-shot learning approaches do not require a meta-training set.  For the VLM-based approaches, we show results using CLIP, MILES, and VideoCLIP by denoting each entry with three values representing the performance of the three VLMs, respectively. We \textbf{bold} the single best accuracy in the table.}
\label{tab:complete-meta-learning-moma-sact}
\end{table}
\section{Hyperparameter selection}
\label{sec:hyperparameter}

To evaluate across all methods consistently, we choose the subset of  hyperparameters outlined below that maximize accuracy on the validation set, and use that same set of hyperparameters for a new run on the training set, reporting resulting accuracy on the test set. We perform this individually for each combination of dataset, VLM backbone, method, and number of shots. \\

\textbf{VL-Prototype.}  For the text weight, we sample 16 values from the interval $[10^{-2}, 10^2]$ log-uniformly.  We also try two prompt ensembles, one is the set of prompts originally used by TIP-Adapter \cite{TIPAdapter}, and the other is the highest performing set of prompts obtained from our own prompt-engineering. We report results averaged across 32 runs for the traditional data splits, and report results averaged across 1,000 runs for the meta-learning splits. \\

\textbf{Linear Probe.} We follow the approach used to run the linear probe baseline in the original CLIP \cite{ImageCLIP} paper and train a logistic regression classifier using L-BFGS with a maximum of 1,000 iterations. We determine $\lambda$, the L2 regularization strength coefficient by doing a hyperparameter sweep on the validation set by sampling 96 values from the interval $[10^{-6}, 10^6]$ log-uniformly.  We report results averaged across 5 runs when using the traditional data splits, and report results averaged across 100 runs for the meta-learning splits. \\

\textbf{CoOp} For CoOp, we search over a context length of 8 and 16, and train for 50 epochs using SGD with batch size of 8.  We perform a grid search over learning rates of $\{6.25\times10^{-5}, 5\times10^{-4}, 2\times10^{-3}, 4\times10^{-3}\}$.  We checkpoint after each epoch and use the checkpoint with the best performance on the validation set. \\

\textbf{Name Tuning.} For Name Tuning, we use "A video of \{\}" as our context prompt. We train for 20 epochs using AdamW with batch size of 8.  We perform a grid search over learning rates of $\{1\times10^{-5}, 1\times10^{-4}, 4\times10^{-3}, 1\times10^{-3}\}$, and $\{0.01, 0.1, 1, 10\}$ for the name regularization hyperparameter, $\alpha$.  We checkpoint after each epoch and use the checkpoint with the best performance on the validation set. \\

\textbf{CoNa.} For CoNa, we use a fixed learning rate of $10^{-5}$, and train for 20 epochs using AdamW with batch size of 8.  We perform grid search over the regularization coefficient $\alpha \in \{1, 5, 10, 20\}$ and use a context length of 4.   We checkpoint after each epoch and use the checkpoint with the best performance on the validation set. \\

\end{document}